\documentclass[runningheads]{llncs}

\usepackage[year=2024,ID=1710]{eccv}

\usepackage{eccvabbrv}
\usepackage{wrapfig}
\usepackage{bbding}
\usepackage{pifont}

\usepackage{graphicx}
\usepackage{booktabs}

\usepackage[accsupp]{axessibility}

\usepackage[pagebackref,breaklinks,colorlinks,citecolor=eccvblue]{hyperref}
\hypersetup{
    urlcolor=magenta
}

\usepackage{orcidlink}

\begin{document}

\title{AnimateZoo: Zero-shot Video Generation of Cross-Species Animation via Subject Alignment} 

\titlerunning{Work in progress}

\author{Yuanfeng Xu\inst{1}\and
Yuhao Chen\inst{1}\and
Zhongzhan Huang\inst{1} \and
Zijian He\inst{1}\and \\
Guangrun Wang\inst{2}\thanks{Corresponding Author.} \and
Philip H.S. Torr\inst{2} \and
Liang Lin\inst{1}}

\authorrunning{Work in progress}

\institute{Sun Yat-sen University \and
University of Oxford\\
\email{\{xuyf93, chenyh387, huangzhzh23, hezj39\}@mail2.sysu.edu.cn, wanggrun@gmail.com, philip.torr@eng.ox.ac.uk, linliang@ieee.org}}

\maketitle

 \begin{figure}[h]
    \centering
    \vspace{-11pt}
   \includegraphics[width=0.95\textwidth]{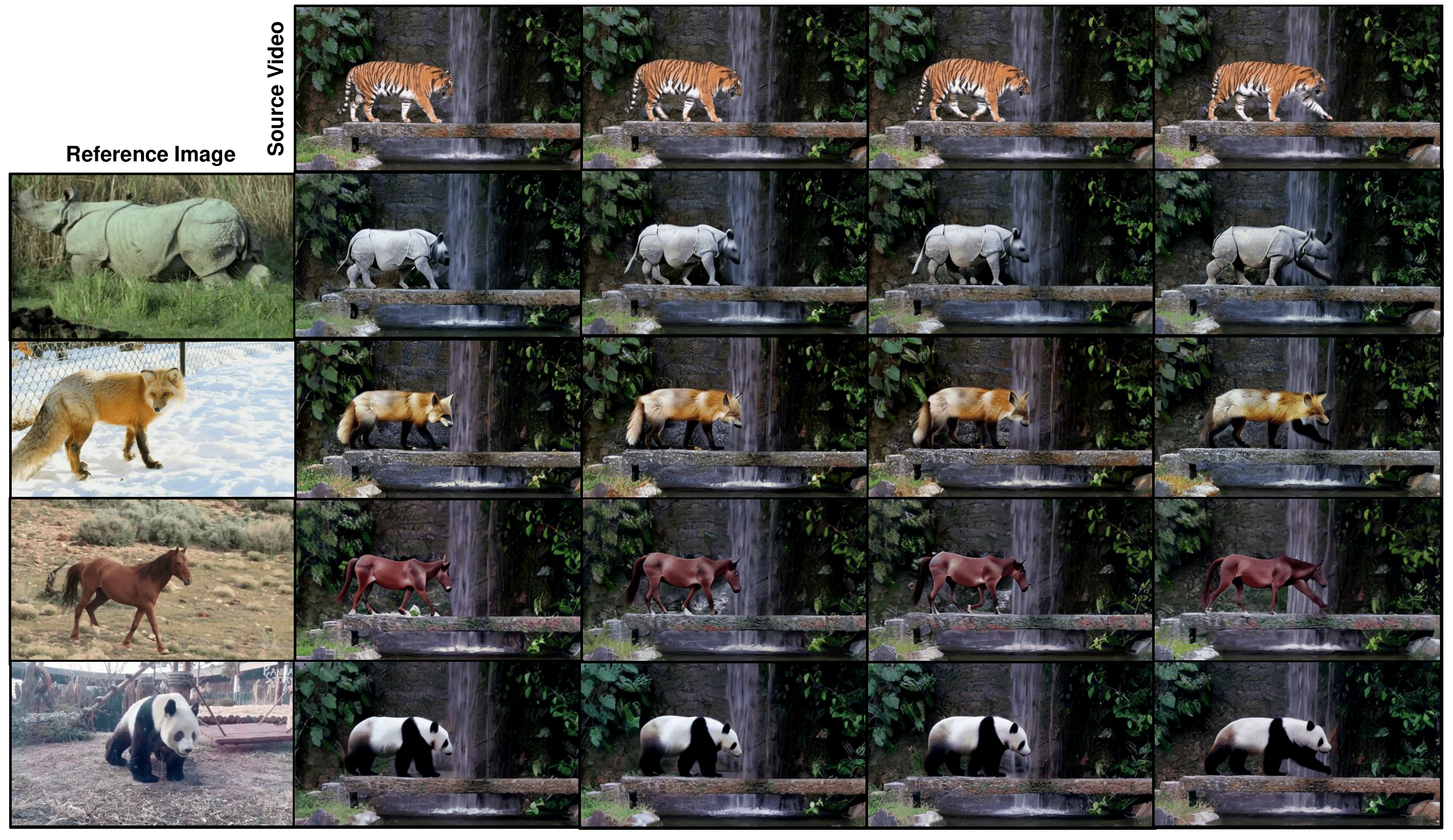}
  \caption{Faithful and controllable  cross-species animation results of \emph{AnimateZoo}. Without any parameter tuning, our model supports seamlessly inheriting actions from diverse animal species while preserving the scene and appearance information consistency.}
  \label{fig:pipeline}
\vspace{-33pt}
\end{figure}

\begin{abstract}

Recent video editing advancements rely on accurate pose sequences to animate subjects. However, these efforts are not suitable for cross-species animation due to pose misalignment between species (for example, the poses of a cat differs greatly from that of a pig due to differences in body structure).
In this paper, we present \emph{AnimateZoo}, a zero-shot diffusion-based video generator to address this challenging cross-species animation issue, aiming to accurately produce animal animations while preserving the background.
The key technique used in our AnimateZoo is subject alignment, which includes two steps. 
First, we improve appearance feature extraction by integrating a Laplacian detail booster and a prompt-tuning identity extractor. These components are specifically designed to capture essential appearance information, including identity and fine details. Second, we align shape features and address conflicts from differing subjects by introducing a scale-information remover. This ensures accurate cross-species animation.
Moreover, we introduce two high-quality animal video datasets featuring a wide variety of species. Trained on these extensive datasets, our model is capable of generating videos characterized by accurate movements, consistent appearance, and high-fidelity frames, without the need for the pre-inference fine-tuning that prior arts required. Extensive experiments showcase the outstanding performance of our method in cross-species action following tasks, demonstrating exceptional shape adaptation capability. The project page is available at \href{https://justinxu0.github.io/AnimateZoo/}{https://justinxu0.github.io/AnimateZoo}.

\end{abstract}

\section{Introduction}
\label{sec:intro}

Animating subjects in static backgrounds by aligning their movements with pose sequences (as in \cite{karras2023dreampose}) or optical flow (as in \cite{peruzzo2024vase}) from other subjects is a crucial task \cite{hu2023animate}.
This task impacts multiple fields such as gaming, filmmaking, and special effects creation, driven by user creativity. By using general skeletal points, virtually any object or animal can be generated. Benefits include lower costs for training animal actors and modeling game NPCs, reduced risks for actors in dangerous scenes, and other advantages.
Prior arts have largely concentrated on \textbf{\emph{intra-species animation}} \cite{xu2023magicanimate,hu2023animate,peruzzo2024vase}, where both the subject in the static image and in the reference video (which could be in the form of pose sequences) belong to the same category, such as both being humans \cite{hu2023animate}. Intra-species animation methods rely on accurate pose sequences to animate subjects. 

Currently, \textbf{\emph{cross-species animation}} remains largely unexplored. The efforts of intra-species animation \cite{peruzzo2024vase, hu2023animate} are not suitable for cross-species animation due to pose misalignment between species. 
Specifically, previous works often struggle to maintain fidelity when main subjects are replaced with subjects from different categories, especially in scenarios involving significant differences (conflicts) in shape, size, or body proportions. Consequently, synthetic videos may exhibit unrealistic postures and inaccurate appearance details.

To tackle the challenge of cross-species animation, we introduce \emph{AnimateZoo}, a diffusion-based framework that facilitates zero-shot video generation using pose sequences from different animals. AnimateZoo is trained just once on a general animal dataset, eliminating the need for pre-inference fine-tuning on specific videos, as previously required. This enables automatic action inheritance from wild animals, broadening the potential applications mentioned above.

AnimateZoo's key technique is appearance alignment between subjects, split into two parts. \textbf{First}, we utilize two feature extractors to gather high-level identity and detailed pixel-level texture information. A domain-specific identity extractor, refined through prompt tuning, captures the animal body's shape and part relationships. Concurrently, a Laplacian detail booster enhances edge and texture details in low-resolution images, requiring minimal computational resources and slight parameter adjustments to preserve fine details. Together, these extractors achieve semantic accuracy and detailed visuals. \textbf{Second}, we introduce a scale-information remover to prevent data leakage during training. By excluding shape and size data, it ensures consistent training and inference, enabling accurate action and shape replication across different animals. This approach guarantees smooth synthesis when using mismatched pose sequences for video generation. In summary, our paper makes four contributions:

\begin{itemize}
    \item We introduce a zero-shot method that inherits actions across animals with varying sizes and body proportions, enabling accurate, high-quality, and cross-species animation, filling crucial gaps in the field.
    \item We design a simplified pipeline for cross-species swapping, integrating a prompt-tuning identity extractor and a Laplacian detail booster for enhanced appearance and texture capture. A scale-information remover guarantees uniform training and inference, reducing shape constraints.
    \item We collect two high-quality animal datasets from existing sources, one on bird movements and another on various animals. Each dataset contains videos of single moving subjects with accurate skeletal annotations and precise foreground masks, rigorously checked by manual inspection, addressing the shortage of high-quality animal video datasets.
    \item Extensive experiments demonstrate that AnimateZoo outperforms prior arts in cross-species animation tasks, showcasing exceptional shape adaptability.

\end{itemize}

\section{Related Works}
\subsection{Video-editing}

Video editing encompasses a spectrum of tasks aimed at altering the visual style\cite{qi2023fatezero}, manipulating object shapes\cite{ho2022imagen,esser2023structure} and motions\cite{wu2023tune}, and more. These techniques typically fall into two overarching categories delineated by their procedural pipelines: explicit editing and implicit editing. Implicit editing approaches leverage warping functions \cite{khachatryan2023text2video} operating within diverse latent spaces to deform reference images or infuse control signals into the generative model, thereby facilitating motion alterations. Conversely, explicit editing methodologies often involve distorting corresponding segments of reference images to match target positions via optical flow information or deformation fields \cite{lee2023shape}. Subsequently, the unaligned content in the target image is supplemented to reconcile any discrepancies with the original image information. Recent research endeavors underscore the efficacy of diffusion-based implicit editing techniques in producing high-fidelity, cohesive, and diverse video content.

\subsection{Data-driven Animation}

Considerable research has focused on animating static images, categorized by methods for subject information extraction and driving mechanisms. Initially, methods\cite{khachatryan2023text2video} relied on textual cues for object movement, providing limited trajectories. Advancements, like ControlNet, integrated diverse modalities for precise control, such as action sequences\cite{hu2023animate}, depth maps\cite{xu2023magicanimate}, and optical flow\cite{peruzzo2024vase}. However, scant attention has been given to utilizing mismatched action information for animating diverse organisms. This presents promising prospects for a universal action control framework to model various animals, ensuring authentic movements across different species and environments.

Prior research has explored diverse methodologies for subject information extraction. While some rely on CLIP image encoders\cite{radford2021learning}, they are limited in capturing intricate image details essential for faithful content generation. Alternatively, methods like Animate Anyone\cite{hu2023animate} and EMO\cite{tian2024emo} employ trainable ReferenceNet architectures to extract detailed visual information via U-Net frameworks\cite{xing2023survey,huang2024scalelong,zhong2023adapter}. However, the complexity of these structures may hinder convergence speed. Moreover, the risk of inundating the model with superfluous details poses challenges, as discussed in Section \ref{sec:Scale_information_Remover}. To address this, we advocate for imposing bottlenecks to manage excessive information influx and selectively removing parts prone to causing information leakage, aligning with our goal of facilitating diverse action generation.

\section{Method}
With subject alignment as the core idea, AnimateZoo aims to facilitate cross-species video subject swapping. Illustrated in \cref{fig:pipeline}, our pipeline utilizes a reference image depicting the desired subject to edit videos describing different animals, seamlessly integrating the target image as the primary character in the synthesized video while retaining the original animals' actions. We first briefly introduce ControlNet and discuss the selection of control information in \cref{sec:Preliminariy}, which dictates the network structure and driving information for our method. \cref{sec:Network_Architecture} provides a detailed explanation of the key components depicted in \cref{fig:pipeline}. \cref{sec:Training_Process} delves into the training process in detail.

\subsection{Preliminariy}
\label{sec:Preliminariy}

\subsubsection{Control Information}
Presently, a variety of methods utilize diverse motion control information, encompassing depth maps, optical flow, point clouds, skeletal points, among others. We compare the results of existing models and our method in the cross-species subject-swapping task in Fig.2. It is evident that existing models are ill-suited for this task, exhibiting sub-optimal performance. Upon analysis, several factors contribute to this limitation. Part of the rationale behind this lies in the mismatch of shape features between the reference subject and action information, a point we will elaborate on in the subsequent \cref{sec:Network_Architecture}. Additionally, overly stringent constraints on control information may exacerbate the issue. CCEdit\cite{feng2023ccedit} relies on depth maps, VASE takes optical flow as driving information. These rigid constraints, particularly challenging for animals in wild environments, often lead to conflicts during training and inference when there exist conflicts in shape, hindering the effective resolution of cross-species video editing tasks. Notably, challenges unique to this task are absent in single-domain tasks like EMO\cite{tian2024emo}, where sound serves as the driving information, or in tasks with minimal human body proportion variations such as virtual try-on, enabling seamless application of the same action sequence to different human photos.

Therefore, our model opts to utilize skeletal points, as opposed to body contour-related information, to drive the video. Unlike optical flow information, skeletal points lack fixed positions, offering scalability during annotation by allowing key points to be distributed within a range. Nevertheless, this flexibility does not imply significant uncertainty in skeletal point-driven videos. On the contrary, our analysis carried out on the bird dataset we proposed demonstrates that well-calibrated skeletal point settings enable precise control. Upon analysis of \cref{fig:short} , it becomes evident that a judicious configuration of skeletal points not only facilitates finer motion control but also enhances the confidence level of generated outcomes. It contributes to the overall efficacy of the generation process. Moreover, inspired by FlowVid, we employ skeletal points as a soft constraint, integrating them with noise maps to initiate the model's denoising process. This setting enhances the adaptability of skeletal points to the subject's shape.

\begin{figure}[tb]
  \centering
  \includegraphics[width=0.95\textwidth]{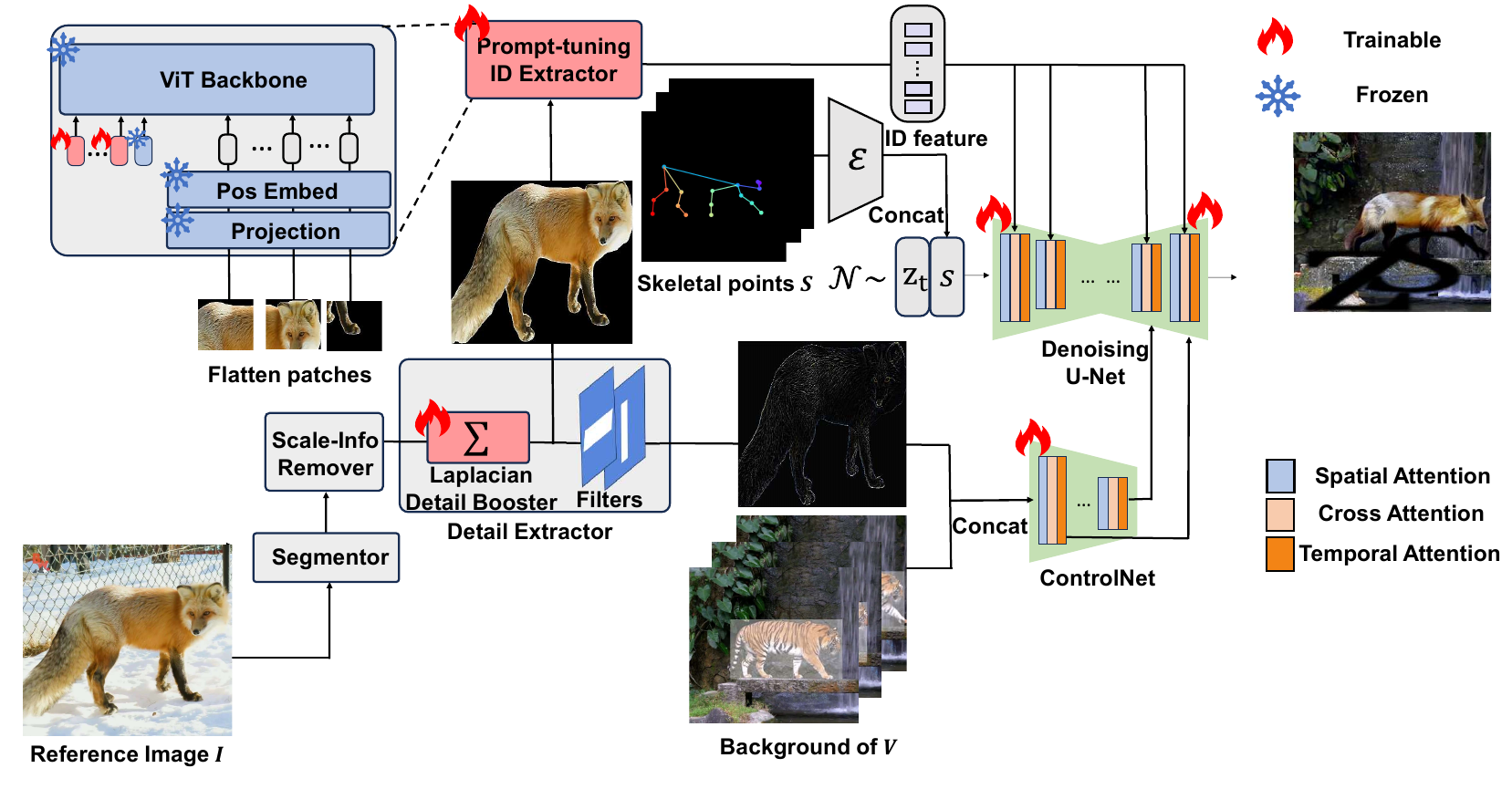}
  \caption{Overview of AnimateZoo, which aims to leverage mismatched pose sequences for driving reference subject motion within a scene. Firstly, we eliminate the background from the reference Image with a segmentation tool. Subsequently, a meticulously crafted scale-information remover removes shape constraints, avoiding conflicts with mismatched skeletal points. Then, we use a Laplacian detail booster to enhance texture and edges in pixel space. Then, high-pass filters are set to preserve pixel-level texture features, which are then concatenated with the video scene slated for editing. Simultaneously, the identity extractor, driven by prompt tuning, extracts appearance features from the enhanced subject. Finally, pose sequence, appearance features, pixel-level textual information, and scene conditions are injected into the diffusion model equipped with temporal structures, facilitating the seamless synthesis result.}
  \label{fig:pipeline}
  \vspace{-11pt}
\end{figure}

\begin{figure}[tb]
  \centering
  \begin{subfigure}{0.45\textwidth}
    \includegraphics[width=0.95\textwidth]{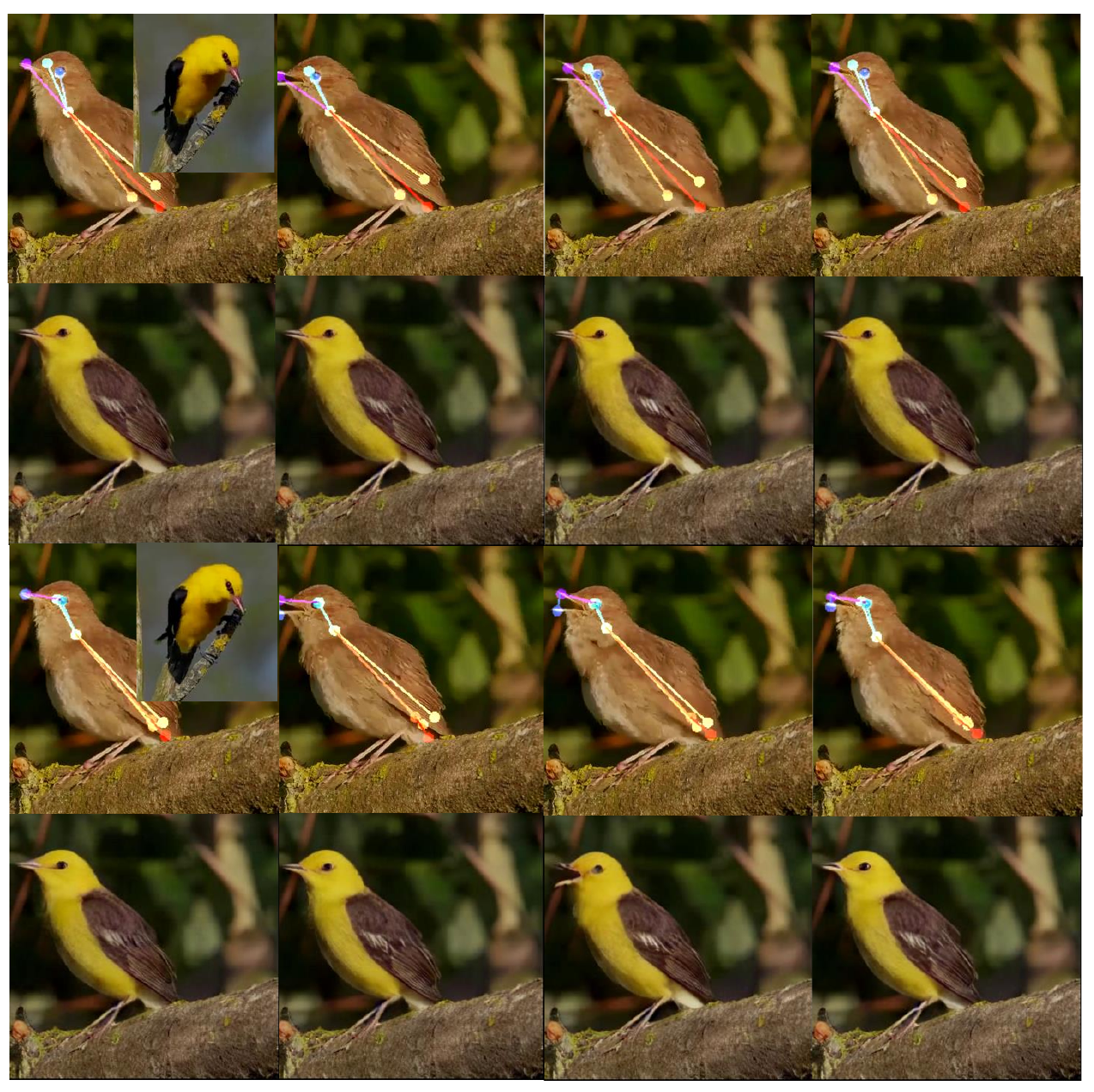}
    \caption{}
    
    \label{fig:bird_1}
  \end{subfigure}
  \hfill
  \begin{subfigure}{0.45\textwidth}
    \includegraphics[width=0.95\textwidth]{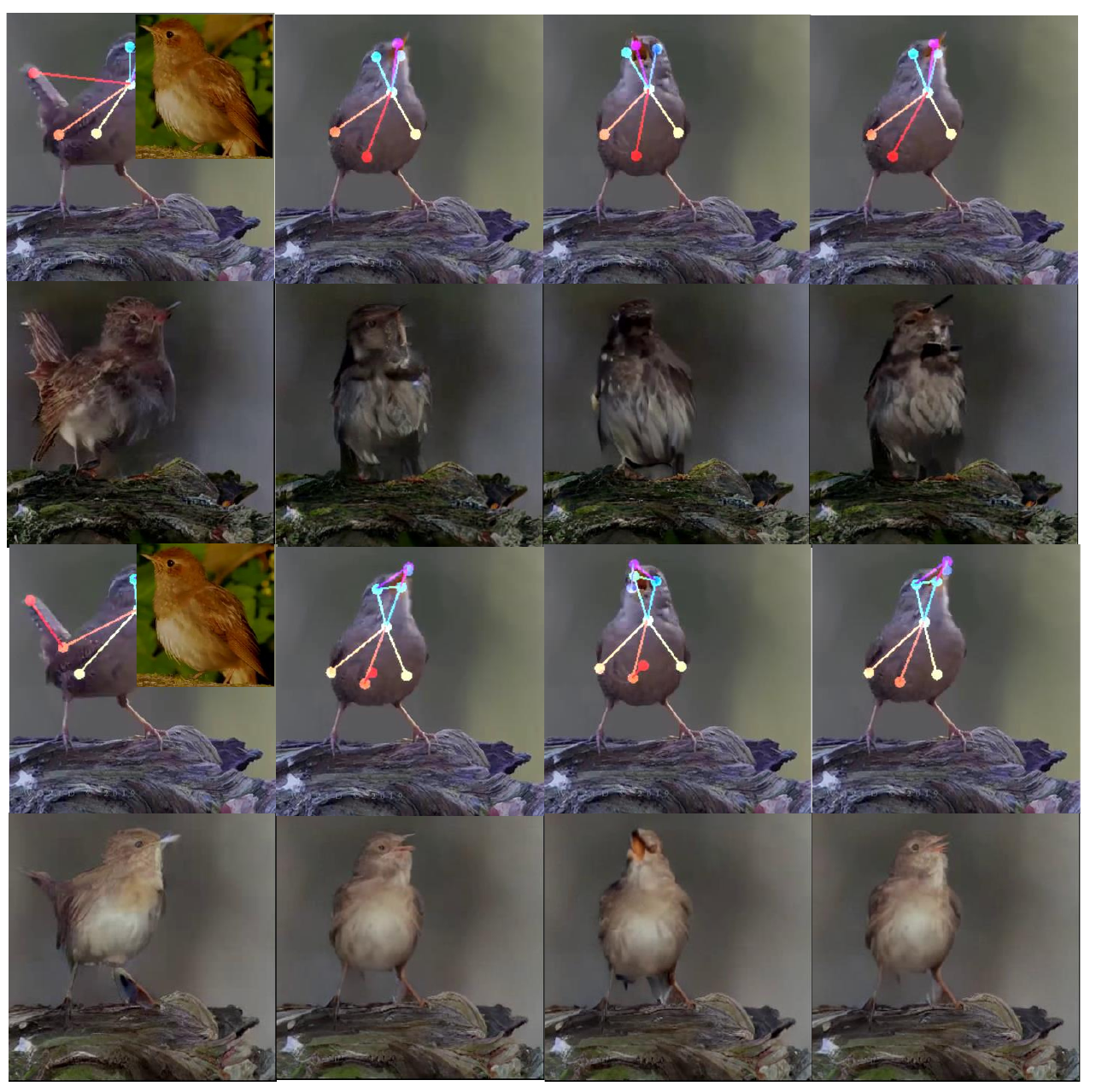}
    \caption{}
    \label{fig:bird_2}
  \end{subfigure}
  \vspace{-11pt}
  \caption{Effect of appropriately positioned skeletal points on motion control. The initial row within each group comprises the video slated for editing, juxtaposed with the reference image situated in the upper-right corner. Subsequently, the ensuing row delineates the outcome of video synthesis. (a) In instances necessitating the depiction of nuanced beak movements, the incorporation of supplementary annotation points facilitates finer manipulation. (b) Conversely, the omission of skeletal points at joints may yield inaccuracies in characterizing the movement of the entire trunk or limb, thereby engendering blurry outcomes. Distinguishing between the abdomen and tail of birds through separate labeling affords superior generation outcomes.}
  \label{fig:short}
  \vspace{-11pt}
\end{figure}

\subsection{Network Architecture}
\label{sec:Network_Architecture}
\subsubsection{Overview.}

Our framework, depicted in Figure \ref{fig:pipeline}, builds upon ControlNet's architecture. It utilizes multiple frames of noise and skeletal point projections in latent space as input, along with detailed information from the reference subject and environmental cues in the control branch. Furthermore, temporal layers are integrated with zero initialization to preserve the generation capability of Stable Diffusion.

To enhance cross-species video subject swapping, our model integrates three key components: (1) Laplacian detail booster, for lightweight fine texture enhancement; (2) Domain-specific identity extractor, using prompt tuning to discern body parts and extract reference subject identity features; (3) Scale information removal, ensuring stable inference during cross-species subject replacement by eliminating partial shape features and scale information of the reference.

\subsubsection{Laplacian Detail Booster.}

Through empirical experimentation, we discovered a crucial insight: feeding higher-resolution foreground images to the identity extraction module significantly enhanced generation quality, while lower-resolution inputs led to unstable shapes and inaccurate textures. This discrepancy arises from the richness of detail in high-resolution images, whereas downsampling can result in information loss, affecting texture and shape accuracy.

However, the down-sampling factor of the identity extractor is fixed. Increasing the input resolution results in generating more patch tokens, significantly expanding memory usage during model execution. Therefore, how to maintain low computational overhead while preserving detailed information becomes the key concern to improve the effectiveness of subject information extraction. 

Inspired by \cite{chen2022estimating,talebi2021learning,tu2023muller}, we adopt a trainable Laplacian resizer, MULLER\cite{tu2023muller}, to boost the detail quality of low-resolution images. 

It combines various filters of distinct frequencies to boost the sample ratio of crucial frequencies for the task at hand, such as edge, detail, and sharpness information. This adjustment involves merely four trainable parameters as weights and biases, rendering the trainable resizer can be applied to our pipeline almost without cost.

\begin{wrapfigure}[18]{r}{0.5\textwidth}
  \centering

  \resizebox{0.99\hsize}{!}{
  \includegraphics[height=6.5cm]{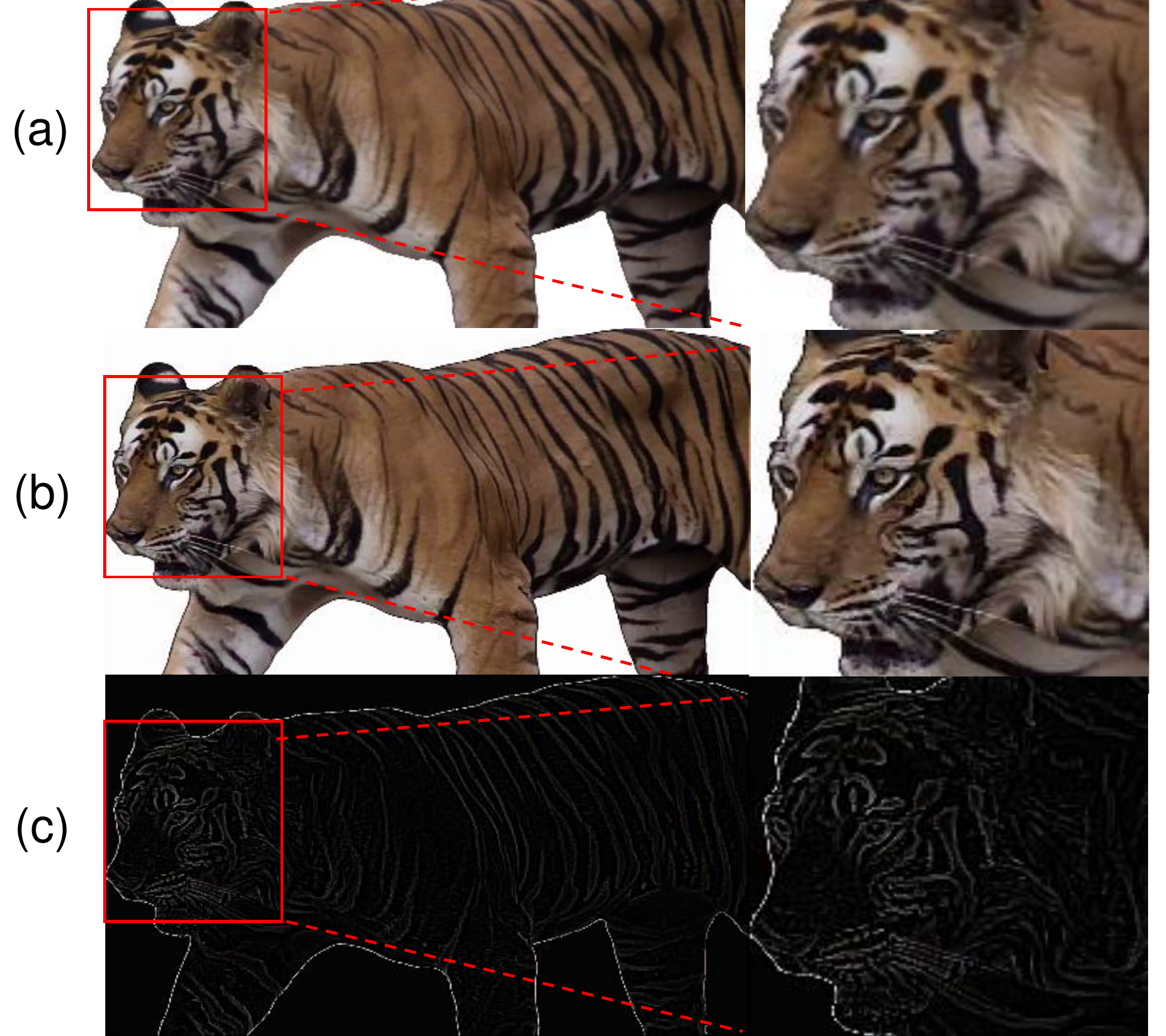}
    
    }
  \vspace{-11pt}
  \caption[]{Comparison of with and without Laplacian detail boosters. (a) Without Laplacian detail booster. (b) With Laplacian detail booster. (c) Difference.}
  \label{fig:tiger}
  \vspace{-11pt}
\end{wrapfigure}

To mitigate the impact of irrelevant background information and improve subsequent component precision, we initially segment the high-resolution reference subject using automatic\cite{kirillov2023segment,qin2020u2} or interactive\cite{chen2021conditional,chen2022estimating,liu2023simpleclick} methods. The segmented subject is then downscaled to low resolution by the Laplacian detail booster, preserving critical visual features and details. With its parameter simplicity and robust regularization, there's no need to modify the original task's loss function, maintaining a task-agnostic structure. Integrating the Laplacian detail booster into the model, as depicted in Figure \ref{fig:tiger}, noticeably enhances various aspects of the tiger's depiction, including fur, beard, and body edges, emphasizing enriched detail and texture. This enhancement highlights the booster's efficacy in accentuating intricate features, thereby improving overall fidelity and realism in edited video content.

\setlength{\intextsep}{1.5pt}
\setlength{\textfloatsep}{1.5pt}

To improve video editing effectiveness, it's crucial to avoid overly restrictive information provision, which can limit diversity and scalability in the generation process. We implement Sobel operators to filter the enhanced reference subject, as validated in prior works\cite{chen2023anydoor}. Horizontal and vertical Sobel operators are applied to process the RGB channels of the foreground image, aiding in line edge and texture detail extraction. Multiplying the result with the original RGB foreground element-wise preserves color details while retaining edge information captures spatial object details.

\subsubsection{Domain-specific identity extractor.}
Unlike commonly used text-guided pre-training models like CLIP, we advocate for models undergoing self-supervised pre-training on large-scale visual data as superior identity information extractors\cite{oquab2023dinov2}. Models trained directly on visual data capture intricate pixel-level details effectively, whereas those trained on text-video pairs may only provide semantic-level supervision, potentially diminishing perception ability for complex visual information.

We select Dinvo-v2 as our base identity information extractor. Pre-trained on extensive image data, Dino-v2 can generate universal visual features and distinguish different parts of various animals, yielding similar visualization results. We believe it is advantageous for controlling animals' actions.

In contrast to previous approaches using pre-trained Dino directly, we employ prompt tuning to dynamically adjust the feature extractor. This strategy reflects our focus on diverse wild animals over generic objects. By integrating a universal visual processor into the training pipeline, we aim to tailor an identity extraction module for specific domains and enhance Dino-v2's ability to capture animals' identity information. All pre-training parameters in the Vision Transformer remain fixed, with only trainable parameters of fixed length concatenated based on the input tokens. This parameter effectiveness learning enables adaptive adjustments in specific task domains while significantly improving training efficiency.

As shown in the magnified view presented in Figure 1, we feed foreground information into the section of the frozen part of Dino with three parts: a frozen class token $T^{1\times768}_{cls}$, patch-by-patch tokens $T^{576\times768}_{patch}$ depending on the reference image, and additional trainable tokens $T^{100\times768}_{learn}$. They are concatenated as the input of the frozen ViT backbone. The output of the extractor, ID feature, is of equal length with inputs. In this way, it allows for efficient adjustment of Dino with minimal trainable units added, mitigating the training overhead and potential performance risks associated with extensive adjustments. The extracted identity information supersedes the position of CLIP information in Stable Diffusion v1.5 and participating in cross-attention calculations.

\subsubsection{Scale-information Remover.}
\label{sec:Scale_information_Remover}

Another empirical finding reveals that superior results are achievable when the target subject and action sequence belong to the same category of animals. However, when they stem from different animal categories, particularly with notable differences in size, body proportions, and posture, issues arise. Despite training on a universal animal dataset, videos generated under such circumstances often exhibit blurred and jittery color blocks around the subject's edges.

This phenomenon arises due to mismatches between the target subject and the action sequence from different sources during inference, which is paired in the training process. Specifically, the skeletal points, which drive the subject's movement, serve as clear indicators of size, posture, and proportion across body parts. However, extracting features directly from the target subject may inadvertently leak prior information, including body size and proportion details, into the model. The patch token from the domain-specific identity extractor incorporates body size and proportion information, and posture prior is contained in some videos that lack significant actions. This may be the reason why lossless information extracted in ReferenceNet performs bad in cross-species task. In summary, using mismatched target animals and action sequences during inference introduces conflicts in shape and posture features, ultimately compromising the model's generation quality.

We address this issue by minimizing shape features and pose priors within the foreground subjects rather than modifying the pose sequences that drive the subjects' motion. To start with, we randomly sample a frame as the reference image from the whole video, alongside randomly selecting the starting point of the video clip to be edited. This technique, also utilized in other studies, bolsters model robustness and serves as a form of data augmentation. Subsequently, we set the segmented foreground subjects to the center of the image to eliminate positional priors. At the same time, we fix the output size of the resizer to regularize the size information of all animals. Finally, random horizontal flipping is adopted to mitigate pose biases inherited from the reference image. These four steps are applied before feature extraction from the foreground information, effectively mitigating inconsistencies between training and inference processes. Consequently, this approach circumvents conflicts stemming from foreground information leakage, facilitating cross-species subject replacement tasks.

\subsection{Training Process}
\label{sec:Training_Process}
\subsubsection{Training Strategy.}

Our training regimen is bifurcated into two distinct stages to effectively harness the potential of the model. Initially, in the first stage, emphasis is placed on leveraging a single video frame to train the entire model, with the exception of the temporal layer. This encompasses the image-diffusion framework, Laplacian detail booster, and identity extractor. This approach accelerates convergence in the spatial dimension, affording the model adept control over the actions of the reference subject. Subsequently, the second stage introduces temporal structures to accommodate multiple frames concurrently, with all other parameters held constant. Training on cohesive video frames facilitates the acquisition of temporal consistency and coherent action generation.

\subsubsection{Model Structure and Initialization.}
Our training regimen comprises two stages to maximize the model's potential. Initially, in the first stage, the focus is on training the entire model, except for the temporal layer, using a single video frame. This includes the image-diffusion framework, Laplacian detail booster, and identity extractor. This approach accelerates convergence in the spatial dimension, granting the model precise control over the actions of the reference subject. In the second stage, temporal structures are introduced to handle multiple frames concurrently, while all other parameters remain unchanged. Training on cohesive video frames fosters the learning of temporal consistency and coherent action generation.

\section{Datasets}
\label{sec:Datasets}
Previous studies underscore the significant impact of datasets on model performance. However, there is currently a dearth of video data specifically curated for animal-related tasks. Moreover, large-scale datasets often suffer from issues such as inconsistent annotation and variable video quality. Training models on such noisy datasets can adversely affect their performance. Datasets related to animal pose estimation exhibit variations in specifications. Table \ref{tab:datasets} provides a comprehensive overview of datasets pertinent to animal pose estimation, with the "Species" column indicating the number of species represented, the "Mask" column denoting the presence of pixel-level masks, and the "Quality" column evaluating occlusion and camera switching prevalent in the video datasets.

Acknowledging the impact of limited high-quality datasets on the progress of animal-related video tasks, we undertake the task of refining and labeling existing open-source animal video datasets, Animal Kingdom and APT. This endeavor involves the creation of dedicated datasets for avian species and comprehensive datasets encompassing 30 distinct animal categories. To foster further research and development in the field, these datasets will be available to the academic community. 

\begin{table}[htbp]
  \centering
  \caption{Datasets related to animal pose estimation.}
    \begin{tabular}{ccccccc}
    \toprule
    Dataset & Type  & \# Frames & Species & Keypoints & Mask & Quality \\
    \midrule
    Animal Pose & Image & 4K    & 5     & $\checkmark$ & $\times$     & \textbackslash{} \\
    AP-10K & Image & 10K   & 54    & $\checkmark$ & $\times$     & \textbackslash{} \\
    \midrule
    Animal Kingdom & Video & 33K   & 1(mainly) & $\times$     & $\times$     & Low \\
    APTv2 & Video & 41K   & 30    & $\checkmark$ & $\times$     & Low \\
    Horse-10 & Video & 8K    & 1     & $\checkmark$ & $\times$     & High \\
    DAVIS (Animals) & Video & 2K    & 15    & $\times$     & $\checkmark$     & High \\
    AnimateZoo-Bird (Ours) & Video & 96K   & 1     & $\checkmark$ & $\checkmark$     & High \\
    AnimateZoo-Universal (Ours) & Video & 53K   & 30    & $\checkmark$ & $\checkmark$     & High \\
    \bottomrule
    \end{tabular}
  \label{tab:datasets}
  \vspace{-11pt}
\end{table}

\subsubsection{Data source.}
The avian dataset presented herein is refined from the Animal Zoo, while the universal dataset is selected from APT. We extensively excluded partial videos with chaotic annotations, instances with overly blurry subjects, or scenes where subjects are excessively obscured. Most of the work during the screening process is done manually.

\subsubsection{Format.}
Both the bird dataset and the universal dataset contain samples depicting the movement of single subjects. The bird dataset consists of 8-frame videos, while the universal dataset comprises 15-frame videos. Each dataset includes the original video, pixel-level mask, and skeletal points of the subject animal. The universal dataset has 17 labeled skeletal points, while the bird dataset has 9. The bird dataset contains 96K frames, while the universal dataset has 53K frames, as detailed in Table \ref{tab:datasets}. Such comprehensive, high-quality datasets within the same animal category are uncommon.

\subsubsection{Processing Methods.}
To facilitate reading and annotation, we initially standardized the length of video clips. 
Regarding the acquisition of skeletal point labels, as APT already includes skeletal points information, we relabel a small subset of samples with inaccurate labeling in the general dataset. For the bird dataset, we initially annotated a small portion of the data and then trained an automatic annotation tool DeepLabCut to automatically label the remaining samples. For obtaining pixel-by-pixel mask annotations, we utilized both automated and interactive tools for processing the two datasets. Bird videos, typically shot using telephoto lenses, often have blurry backgrounds. Automated salient object detection tools can easily annotate the outlines. Conversely, the background of examples in universal datasets tends to be more complex and varied. We employed interactive annotation tools to manually gather mask information for universal animals. When there are multiple subjects in the camera, it is common to annotate the original video separately after duplicating it.

\section{Experiments}

\subsection{Implementation Details}

\subsubsection{Training Settings.}
Training video clips are extracted from a long video, beginning with a randomly determined start frame, and subsequently preserving a consecutive sequence of frames. The reference frame for each clip is randomly sampled from the entire video. The length of training video clips is set to 1 for the first stage, which is subsequently extended to 8 for the second stage. Each frame is resized to a resolution of 256 $\times$ 448, and no additional data augmentation operations, such as random flip or center cropping, are applied. The model runs 15,000 iterations with a batch size of 64 in the first stage, while in the second stage, it iterates 10,000 steps with a batch size of 16. AdamW is employed as the optimizer across both stages, with an initial learning rate of 1e-5. Experiments are conducted on a single NVIDIA A40 GPU.

\subsubsection{Benchmarks.}
To showcase the efficacy of our approach for a diverse range of actions, we choose animals in the wild for video editing. To demonstrate the robustness of our method for accommodating subjects with varying sizes and body proportions, we select the dataset we proposed to compare performance with other methods. Spanning 30 distinct categories of animals, it provides a wide evaluation of the adaptability across varied subjects. 

\subsubsection{Metrics.}
In assessing the efficacy of our model relative to alternative approaches, we employ a comprehensive suite of metrics to evaluate the generated results. Firstly, we evaluate the average quality of individual frames utilizing standard metrics including SSIM, PSNR, and LPIPS. Additionally, we leverage CLIP-T to gauge temporal consistency to assess the similarity between edited frames and reference subjects. Moreover, we conducted a user study in \cref{sec:user_study} involving 30 volunteers to provide subjective evaluations of the generated results, focusing on aspects including fidelity, quality, and temporal coherence.

\subsection{Comparisons}

\begin{figure}[tb]
  \centering
  \includegraphics[width=1.0\textwidth]{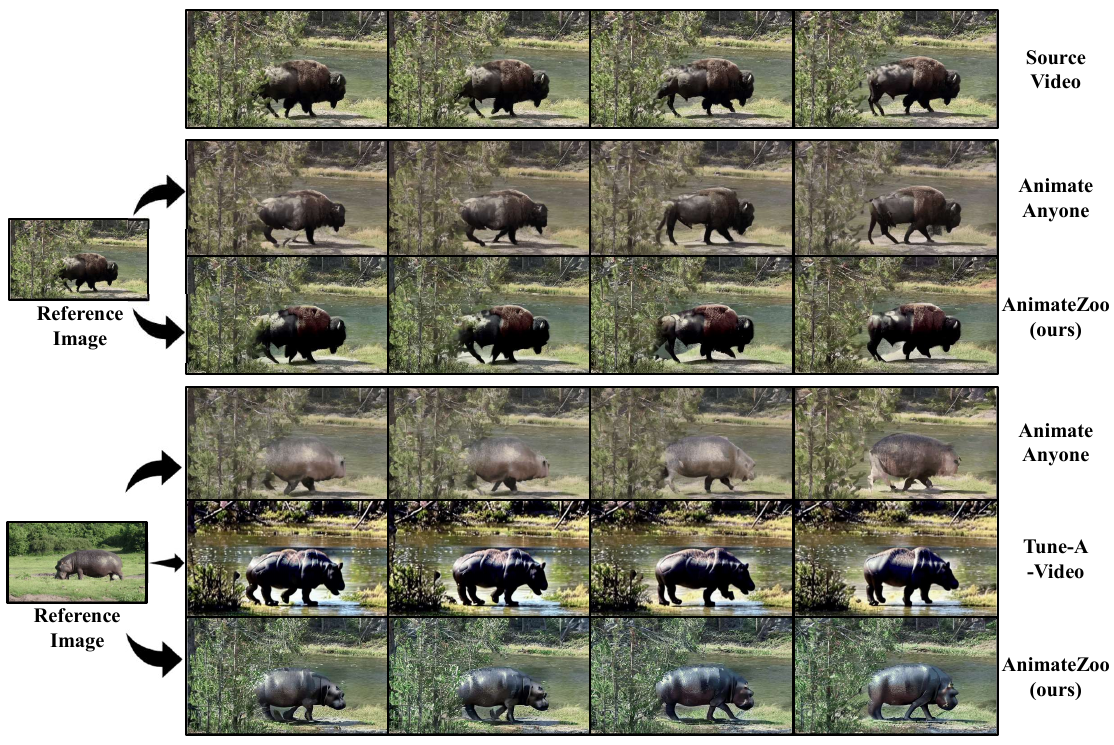}
  \caption{Visualizations of AnimateZoo and other advanced methods across various tasks are presented herein.}
  \label{fig:exp}
\end{figure}
In this section, we assess the efficacy of our model across two distinct tasks.
(1) Animation within a specific domain, which entails animating a designated subject within the background of a reference image based on precise pose sequences. This task serves as a standard evaluation benchmark within the field.
(2) Cross-species animation involves driving the movements of different animals using mismatched pose sequences from distinct species. The varying body structures among different species often result in unstable motion and jittery artifacts when employing mismatched bone point-driven video generation techniques. This phenomenon results in blurred and degraded visual outcomes.

By relaxing constraints on foreground shape information, our method ensures consistency between training and sampling processes, preserving fidelity even for subjects with significant body proportion differences. Despite training on well-matched skeletal points-subject pairs, our model excels in preserving subject fidelity during inference. Leveraging this capability, we benchmark our model against other video editing techniques across various tasks.

Currently, several video editing methods offer direct video generation tailored to specific actions without parameter tuning. Our model stands out among these. While Animate Anyone has shown impressive performance in human animation tasks, its background modification capabilities are limited, relying solely on a reference image without replacing backgrounds from other videos. To ensure fair comparisons, we concatenated background and action sequences as inputs for Denoising U-Net, training two models with identical settings.

We provide visualizations of three methods in Figure \ref{fig:exp}, where the first row shows the source video with the reference image. The subsequent two rows depict the generation effects of Animate Anyone and AnimateZoo, driven by the identity information extracted from the source video. Both Animate Anyone and our method demonstrate commendable performance, highlighting the efficacy of both approaches.

The last three lines of the \cref{fig:exp} depict video content generated using mismatched subject and action information based on Animate Anyone, Tune-a video, and our method. Notably, significant deformations are observed in the animated animation results, leading to actions inconsistent with the original video, highlighting challenges in resolving conflicts in body structure between the reference subject and the original video. In contrast, utilizing BLIP2 as the source of text prompt in the fifth line of the image, Tune-a-video produces reasonable results. However, due to inherent limitations in representing complex visual information through textual prompts, a notable gap persists between the generated results and the reference image. Conversely, our design circumvents potential structural conflicts between the subject and the original video by introducing scale information removal and refining promotion information from the reference subject. Instead, the model extracts shape information solely from keypoints, ensuring compatibility and coherence between the subject and the original video. This design choice safeguards the generation of high-quality videos with minimal risk of structural inconsistency

\begin{table}[htbp]
  \centering
  \caption{Quantitative comparison with other advanced methods.}
    \begin{tabular}{cccccc}
    \toprule
          & DOVER($\uparrow$) & SSIM($\uparrow$)  & PSNR($\uparrow$)  & LPIPS($\downarrow$) & CLIP-T($\uparrow$) \\
    \midrule
    Tune-A-Video & 45.85 &  \textbackslash{}     &   \textbackslash{}    &  \textbackslash{}     & \textbf{0.9919} \\
    Animate Anyone & \textbf{50.48} & 0.43  & 15.12 & 0.1712 & 0.9456 \\
    Ours  & 49.39 & \textbf{0.7964} & \textbf{22.43} & \textbf{0.1246} & 0.9505 \\
    \bottomrule
    \end{tabular}
  \label{tab:addlabel}
\end{table}
The quantitative analysis results are presented in Table 1. It is evident that our method outperforms Animal Anyone in terms of SSIM, PSNR and LPIPS, indicating the superior accuracy of our generated actions. Additionally, our CLIP-T score slightly surpasses that of Animal Anyone, albeit our aesthetic score is marginally lower than the DOVER score. This observation suggests that our Laplacian detail booster captures more intricate features, potentially leading to a slightly reduced aesthetic score due to uneven shapes. Moreover, in the absence of ground truth in Tune-a-video, only DOVER and CLIP-T indicators can be computed. Comparing our DOVER score with that of Tune-a-video, we observe a higher score, while the CLIP-T score is lower. This disparity suggests that our fine-tuning-based method exhibits superior performance in terms of temporal coherence.

\begin{figure}[tb]
  \centering
  \includegraphics[width=\textwidth]{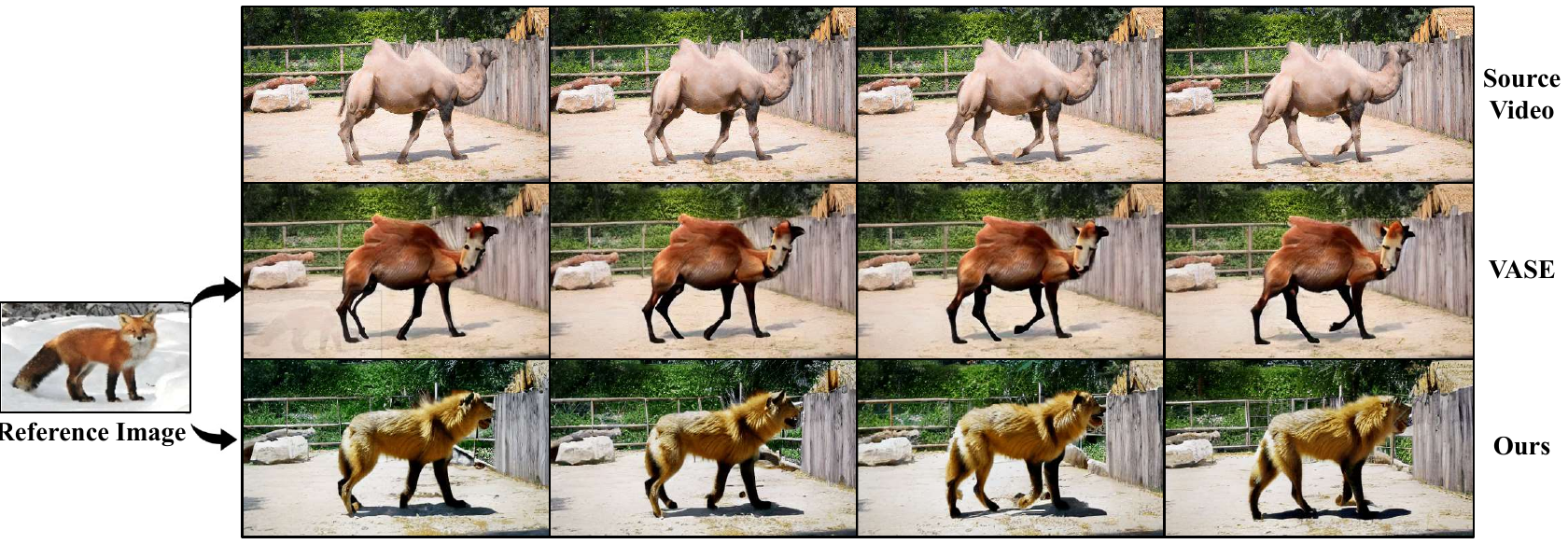}
  \caption{Visualizations of AnimateZoo and VASE\cite{peruzzo2024vase} on DAVIS.}
  \label{fig:davis}
\end{figure}
Unlike the zero-shot method, VASE relies on optical flow and extracting warping relationships to manipulate the subject in the source video to achieve the target subject. This approach necessitates adjustments to the parameters of both the source video and the reference subject prior to each inference. To ensure fairness, we conducted performance comparisons on another benchmark, DAVIS. Observing the results depicted in \cref{fig:davis}, it becomes apparent that VASE can only effect minor modifications to the original subject's shape, such as smoothing out camera bumps, filling in buttons, and making unreasonable adjustments to head posture. In contrast, our model demonstrates superior performance in accurately simulating animal movements and faithfully preserving the details of the subject's shape.

\subsection{Ablation Studies}

\begin{table}[htbp]
  \centering
  \caption{Quantitative analysis elucidates the distinct contributions of various components within the model.}
    \begin{tabular}{cc|ccc}
    \toprule
    Laplacian Detail Booster & Prompt Tuning Extractor & SSIM($\uparrow$)  & LPIPS($\downarrow$) & CLIP-T($\uparrow$) \\
    \midrule
    $\times$     & $\times$     & 0.793 & 0.1248 & 0.9588 \\
    $\checkmark$   & $\times$     & 0.796 & 0.1246 & 0.9596 \\
    $\times$     & $\checkmark$   & 0.755 & 0.136 & 0.9605 \\
    $\checkmark$   & $\checkmark$   & \textbf{0.7964} & \textbf{0.1246} & \textbf{0.9626} \\
    \bottomrule
    \end{tabular}
  \label{tab:addlabel}
\end{table}

Extensive ablation experiments were undertaken, featuring various design variants to assess the efficacy of our approaches. As depicted in Table \ref{tab:addlabel}, components utilizing prompt tuning, particularly CLIP-T, achieve higher scores and demonstrate superior temporal consistency. This underscores the effectiveness of prompt tuning operations in enabling adaptive adjustments for the domain, consequently enhancing the extraction of stable appearance features and bolstering temporal consistency.

\section{Conclusion}

In this study, we introduce AnimateZoo, a diffusion-based video generator enabling cross-species action control while addressing shape feature differences among subjects. Our key innovation lies in achieving subject alignment through three components: utilizing complementary elements such as the Laplacian detail booster and prompt tuning identity extractor for shape information extraction, and integrating a scale information remover to prevent shape information leakage. This ensures training and testing consistency, facilitating precise action control and fidelity in video generation. Additionally, our method provides a novel solution for universal video subject control.

\bibliographystyle{splncs04}
\bibliography{main}

\newpage
\appendix
\begin{center}
      {\bf APPENDIX}
    \end{center}
We elaborate on our method's additional results and analysis in the appendix. Specifically, \cref{sec:Additional_results} presents more cases of our method. Then, we illustrate another application: generating animals within a blank video scene and controlling their motions. \cref{sec:Comparison} includes a comparison with advanced methods\cite{hu2023animate,wu2023tune}, focusing on generation efficiency and time consumption, along with a user study assessing video coherence, background and subject fidelity, and visual quality. Finally, we detail the limitations of existing models in \cref{sec:Limitations}.

\begin{figure}[h]
  \centering
  \includegraphics[width=0.95\textwidth]{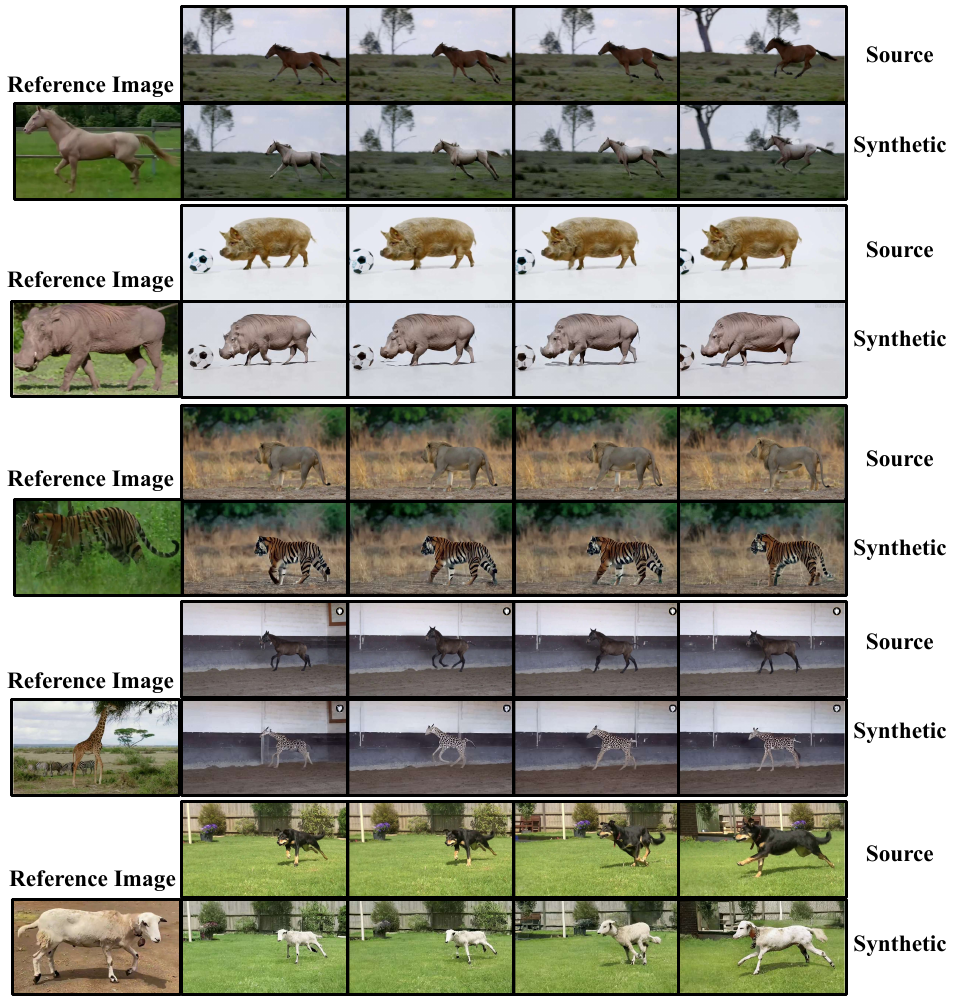}
  \caption{More cases of AnimateZoo.}
  \label{fig:other_cases}
\end{figure}

\section{Additional results}
\label{sec:Additional_results}
\subsection{More Cases in Cross-Species Animation}
Due to space constraints, we've presented a few cases in the main text. To showcase the outstanding performance of our model across various scenarios, we present additional cases here, as depicted in \cref{fig:other_cases}. 

\cref{fig:other_cases} illustrates AnimateZoo's efficacy in diverse scenarios with 5 sets of cases. Each case comprises two rows: the first row depicts the source video, while the second row showcases synthetic frames. The leftmost image in the second row serves as the reference image. Whether animating subjects within the same species (the first and second groups) or conducting cross-species animation (the third, fourth, and fifth groups), AnimateZoo consistently generates reasonable results.

\subsection{Another Application: Creating Something Out of Nothing}

\begin{figure}[h]
  \centering
  \includegraphics[width=0.95\textwidth]{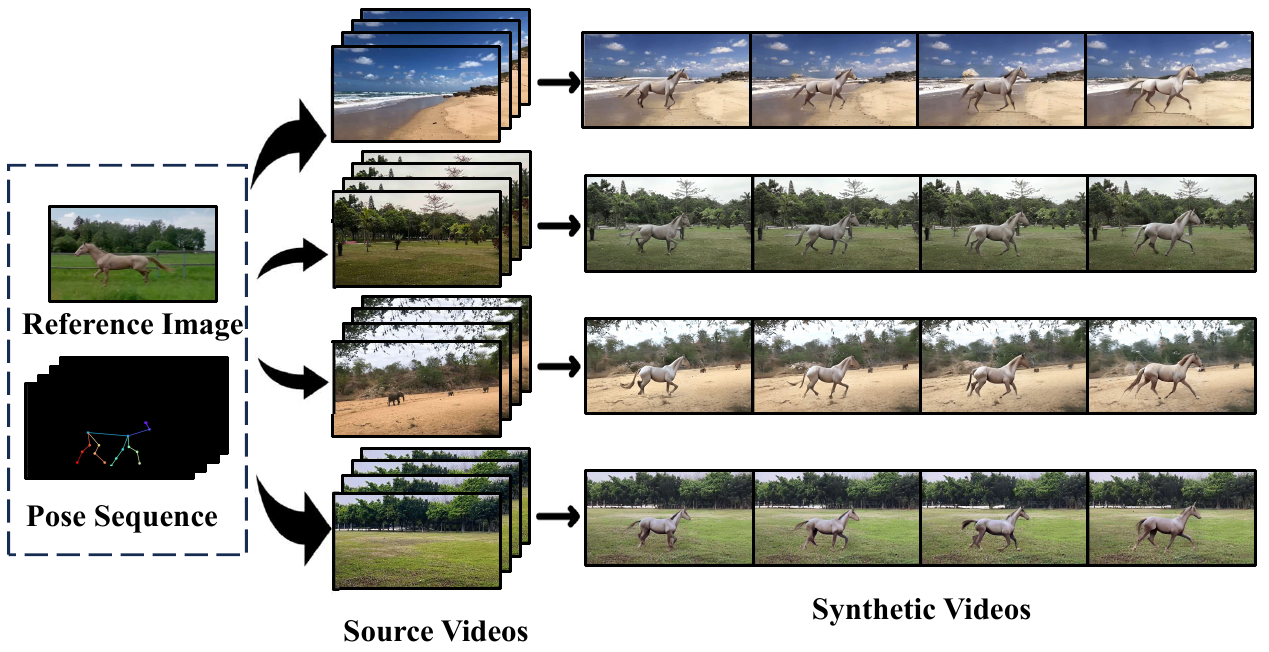}
  \caption{Visualizations of subject transplantation across diverse scenarios.}
  \label{fig:same_subject}
\end{figure}

\begin{figure}[h]
  \centering
  \includegraphics[width=0.95\textwidth]{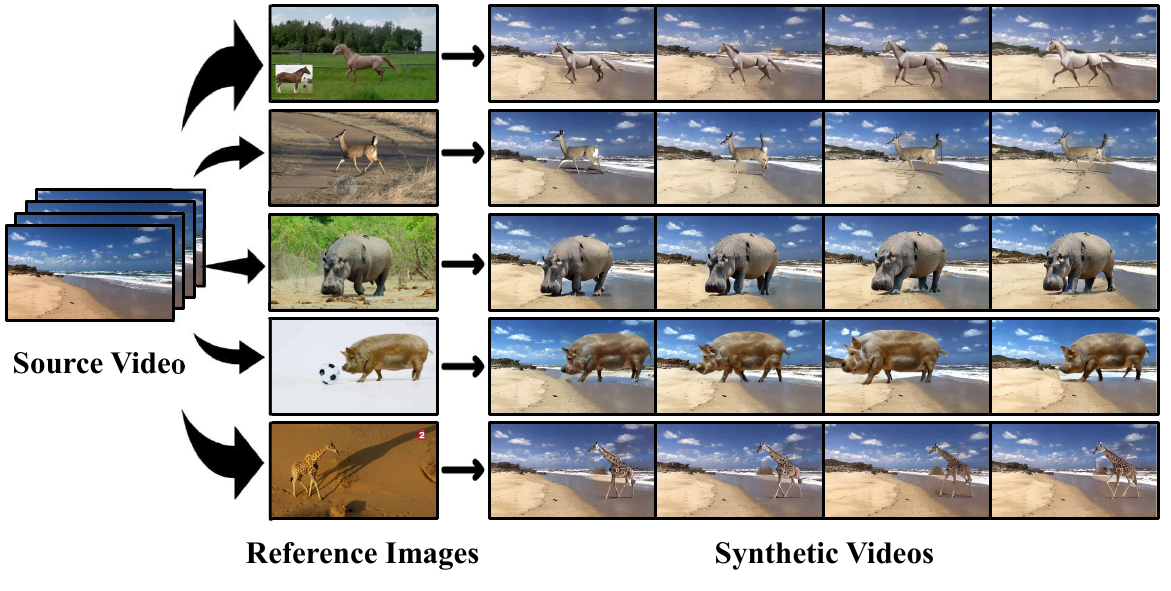}
  \caption{Visualizations of subject transplantation with diverse subjects.}
  \label{fig:same_backgrd}
\end{figure}

In contrast to prior instances where animals were replaced in the original videos, we explored another application of AnimateZoo: transplanting given subjects to open-field backgrounds and applying matched or mismatched pose sequences for controlling. \cref{fig:same_subject} illustrates this by transplanting the same subject and action across various scenarios, while \cref{fig:same_backgrd} showcases different subjects and actions in the same scenario.

\cref{fig:same_subject} is organized into four groups, each sharing the same theme and action sequence, as denoted within the dashed box on the left. These groups showcase videos synthesized from various source videos. In \cref{fig:same_backgrd}, five groups of examples maintain a consistent background while utilizing different reference images for video synthesis. Both figures demonstrate AnimateZoo's proficiency in transplanting various animals into diverse scenes, offering an intriguing avenue for applications in movie effects and game modeling despite its current focus solely on animals.

Notably, the backgrounds featured in \cref{fig:same_subject} and \cref{fig:same_backgrd} are sourced from external datasets or manually captured videos, underscoring the generalization of our method with real-world video content.

\section{Comparison with Advanced Methods}
\label{sec:Comparison}

\subsection{User Study}

\label{sec:user_study}
To thoroughly assess video generation quality across AnimateZoo, Tune-A-Video\cite{wu2023tune}, and Animate Anyone\cite{hu2023animate}, we devised rating criteria encompassing quality, background fidelity, subject fidelity, and coherence. Quality gauges realism and coordination in generated videos, while background fidelity measures alignment with source video backgrounds. Subject fidelity evaluates resemblance to reference subjects, and temporal coherence assesses frame-to-frame consistency. We engaged 36 annotators to rate 3 video groups, each comprising source videos, the reference image, and generated results from the three methods. Notably, Tune-A-Video\cite{wu2023tune} operates on textual prompts. To maintain objectivity, we utilize BLIP2 to extract subject descriptions and then fine-tune parameters accordingly. Regarding the scoring indicators, we follow the method used in \cite{chen2023anydoor}, where annotators sort each metric and assign scores of 3, 2, and 1 to each item from top to bottom, with average scores tabulated in \cref{tab:user_study}.

\begin{table}[htbp]
  \centering
  \caption{User study on AnimateZoo and other advanced works. Each scoring item is rated from 1 (worst) to 3 (best).}
    \begin{tabular}{ccccccccccccccccccccccc}
    \toprule
          &       &       & Quality$(\uparrow)$ & Subject Fidelity$(\uparrow)$ & Background Fidelity$(\uparrow)$ & Coherence$(\uparrow)$ \\
    \midrule
    Tune-A-Video &       &       & 2.03  & 1.95  & 1.76  & 1.89 \\
    Animate Anyone &       &       & 1.44  & 1.56  & 2.06  & 1.68 \\
    AnimateZoo (ours) &       &       & \textbf{2.53} & \textbf{2.49} & \textbf{2.19} & \textbf{2.44} \\
    \bottomrule
    \end{tabular}%
  \label{tab:user_study}%
\end{table}%

It can be observed that our method demonstrates superior performance across all metrics compared to existing methods, particularly excelling in subject fidelity. This underscores the efficacy of our approach in achieving robust subject alignment.

\begin{figure}[h]
  \centering
  \includegraphics[width=0.95\textwidth]{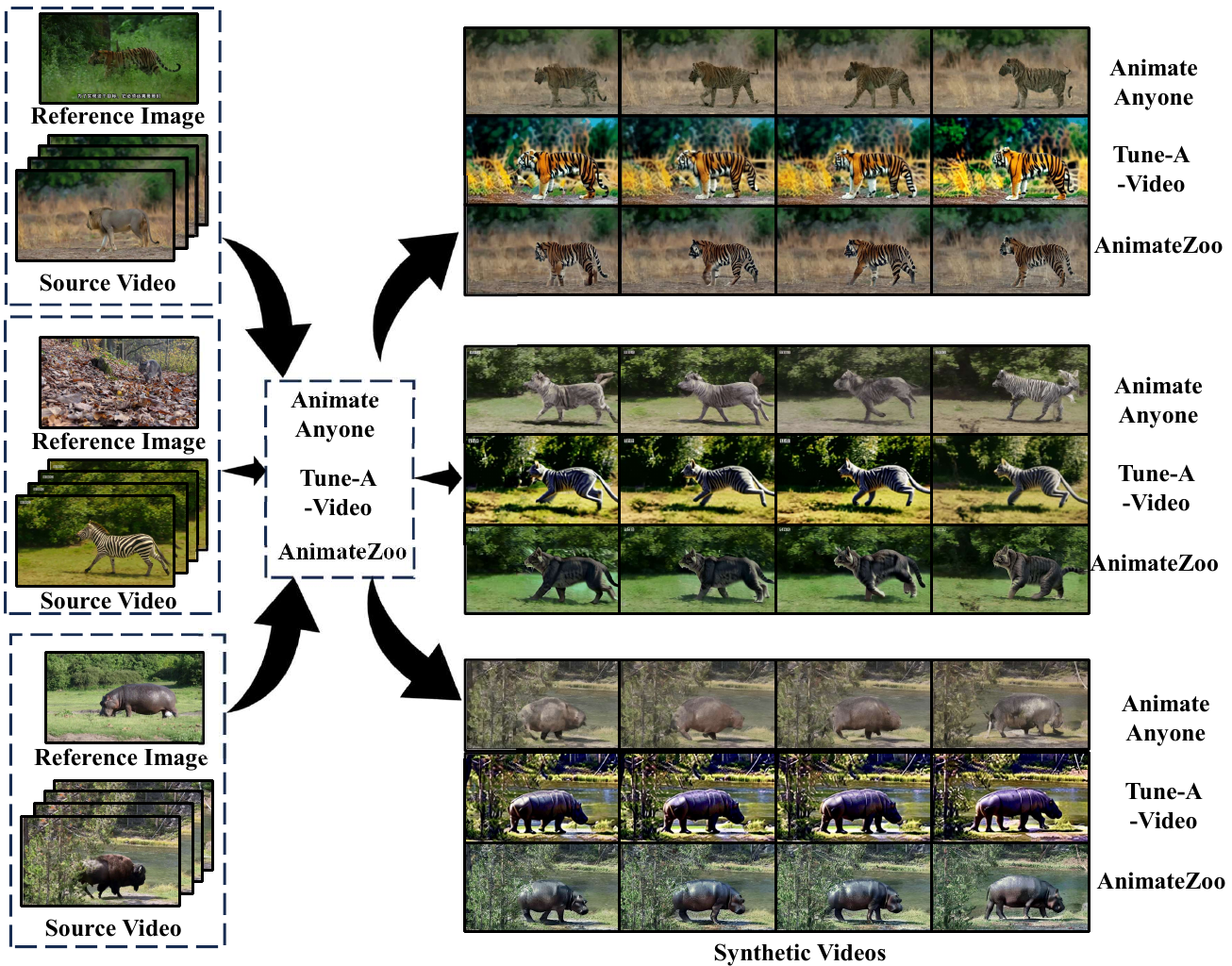}
  \caption{Visualizations used in User Study.}
  \label{fig:user_study_figure}
\end{figure}

The images and videos utilized in User Study are depicted in \cref{fig:user_study_figure}. In each comparison group, three methods are employed for video synthesis based on the source video and reference image on the left: Animate Anyone, Tune-A-Video, and AnimateZoo. The resulting videos were displayed on the right side of the image. AnimateZoo produces superior videos across all evaluated aspects.

\subsection{Time Efficiency}

\begin{figure}[tb]
  \centering
  \includegraphics[width=0.8\textwidth]{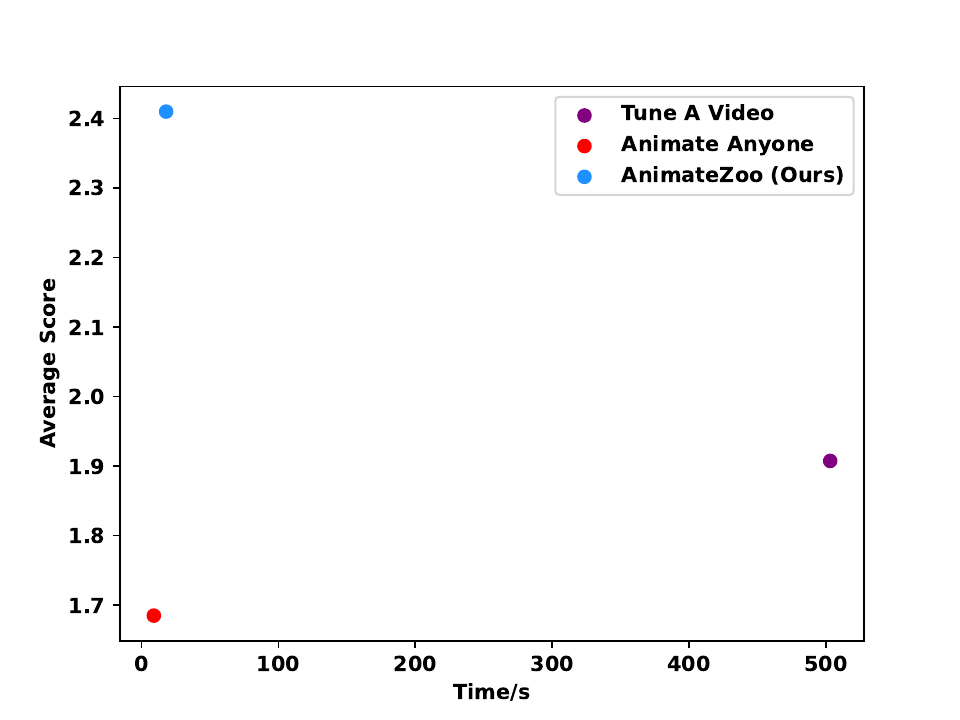}
  \caption{A comparison of time consumption and performance across various methods.}
  \label{fig:time_comparison}
\end{figure}

Among the methods evaluated in the User Study, Animate Anyone and AnimateZoo operate as zero-shot methods, whereas the Tune-A-Video method necessitates parameter tuning before each inference. To provide a clearer insight into the correlation between model performance and time consumption, \cref{fig:time_comparison} plots the total inference time against the average scores of the four metrics listed in \cref{tab:user_study}. It's evident that our method achieves a faster inference process while maintaining higher video quality, distinguishing it from the other methods.

\section{Limitations and Future Work}
\label{sec:Limitations}

\begin{figure}[tb]
  \centering
  \includegraphics[width=0.95\textwidth]{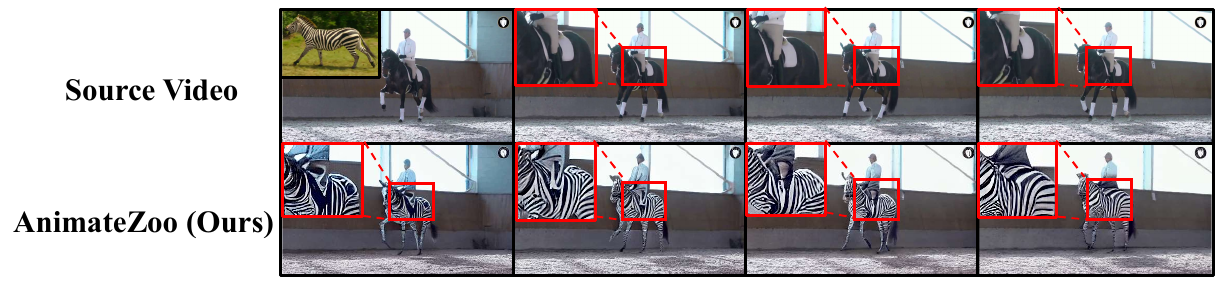}
  \caption{
Limitations: Our method may obscure content at interaction sites involving the controlled subject and other animals.}
  \label{fig:limitaions}
\end{figure}

While AnimateZoo successfully achieves cross-species animation, there exist certain limitations in our current implementation. \cref{fig:limitaions} illustrates a failure case where our method struggles with interactions between multiple objects in the video. The first row displays the source video, followed by the generated results from AnimateZoo in the second row. The upper left corner of the first image in the first row serves as the reference image, with the remaining images in that corner being enlargements of the human leg. Specifically, when confronted with occlusion caused by human legs on the zebra's body, AnimateZoo erroneously fills the occluded region with texture from the zebra, neglecting the presence of human legs. This flawed generation process may stem from insufficient attention to close interactions between animals within the training dataset. A potential solution could involve augmenting the dataset with samples that focus on animal interactions, followed by fine-tuning or retraining the model. This avenue for improvement is earmarked for future in-depth research efforts.

\end{document}